\documentclass[conference,compsoc]{IEEEtran}
\usepackage[bookmarks=false]{hyperref}
\usepackage{amsmath}
\usepackage{hyperref}
\usepackage{url}
\usepackage[nocompress]{cite}
\usepackage[pdftex]{graphicx}
\usepackage{enumitem}
\usepackage{algorithm}
\usepackage[noend]{algpseudocode}
\usepackage{multicol}
\usepackage{caption} 
\usepackage{array}
\usepackage[caption=false,font=footnotesize,labelfont=sf,textfont=sf]{subfig}
\usepackage{fixltx2e}
\usepackage{stfloats}

\newlist{inlinelist}{enumerate*}{1}
\setlist*[inlinelist,1]{%
  label=(\roman*),
}

\begin{document}

\title{Dataset Construction via Attention for Aspect Term Extraction \\ with Distant Supervision}

\author{\IEEEauthorblockN{Athanasios Giannakopoulos\IEEEauthorrefmark{1}\IEEEauthorrefmark{2}\IEEEauthorrefmark{3}, Diego Antognini\IEEEauthorrefmark{2}\IEEEauthorrefmark{3}, Claudiu Musat\IEEEauthorrefmark{1}, Andreea Hossmann\IEEEauthorrefmark{1} and Michael Baeriswyl\IEEEauthorrefmark{1}}
\IEEEauthorblockA{\IEEEauthorrefmark{1}Artificial Intelligence Group~|~Swisscom AG\\
Email: \texttt{\{firstName.lastName\}@swisscom.com}}
\IEEEauthorblockA{\IEEEauthorrefmark{2}Artificial Intelligence Laboratory~|~EPFL, Lausanne, Switzerland\\
Email: \texttt{\{firstName.lastName\}@epfl.ch}}
\IEEEauthorblockA{\IEEEauthorrefmark{3}Equal contribution}
}

\maketitle

\begin{abstract}
Aspect Term Extraction (ATE) detects opinionated aspect terms in sentences or text spans, with the end goal of performing aspect-based sentiment analysis. The small amount of available datasets for supervised ATE and the fact that they cover only a few domains raise the need for exploiting other data sources in new and creative ways. Publicly available review corpora contain a plethora of opinionated aspect terms and cover a larger domain spectrum. In this paper, we first propose a method for using such review corpora for creating a new dataset for ATE. Our method relies on an attention mechanism to select sentences that have a high likelihood of containing actual opinionated aspects. We thus improve the quality of the extracted aspects. We then use the constructed dataset to train a model and perform ATE with distant supervision. By evaluating on human annotated datasets, we prove that our method achieves a significantly improved performance over various unsupervised and supervised baselines. Finally, we prove that sentence selection matters when it comes to creating new datasets for ATE. Specifically, we show that, using a set of selected sentences leads to higher ATE performance compared to using the whole sentence set. 
\end{abstract} 

\section{Introduction} \label{sec:introduction}
The majority of current sentiment analysis approaches focuses on detecting the overall polarity of a sentence or text span. However, the overall polarity refers to a broader context, instead of identifying specific targets. In addition, many sentences or paragraphs contain both positive and negative polarities, which complicates the assignment of a correct overall polarity.

Aspect based sentiment analysis (ABSA) is a more detailed and in-depth approach compared to traditional sentiment analysis and aims at tackling the problems of the latter. ABSA can be decomposed in two tasks: 
\begin{enumerate}
    \item Aspect term extraction (ATE), where the goal is to identify all aspect terms (e.g. battery, screen) of the target entity (e.g. laptop) in a sentence (or text span). 
    \item Sentiment Polarity (SP), where the goal is to identify the polarity (e.g. positive) attached to each aspect term.
\end{enumerate}

We focus on ATE (rather than SP), because it is a harder and more interesting problem. The existing learning techniques for ATE can be categorized into supervised and unsupervised. Each category comes with numerous benefits and drawbacks. Supervised ATE leads to high performance on unseen data. However, the available human annotated datasets are restricted to only a few domains (e.g. restaurants and laptops) and are very small. Even the biggest available human annotated datasets~|~provided by the SemEval ABSA contest~|~contain only a few thousand sentences. Unsupervised ATE overcomes the aforementioned problems by exploiting large and publicly available opinion texts, such as review corpora. These corpora cover a larger domain spectrum (e.g. books, food, electronic devices, etc.) and allow us to perform ATE  in a domain-independent fashion. However, unsupervised systems for ATE come at the cost of lower performance compared to supervised ones~\cite{semeval2014}.

We propose a third option, namely ATE with distant supervision. To the best of our knowledge, there is no prior work in this area. To this end, we first propose a novel attention-based method to construct datasets for ATE, starting from review corpora, which are naturally rich in opinionated aspect terms. Using the constructed dataset, we introduce a new model for ATE, which performs feature extraction and aspect term detection simultaneously while training.

\begin{figure}[t]
    \begin{center}
    \includegraphics[width=0.48\textwidth]{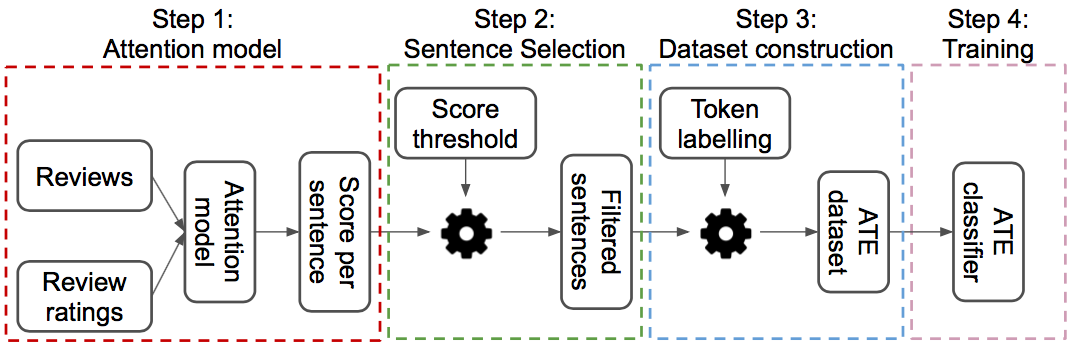}
    \end{center}
    \caption{Pipeline for ATE with distant supervision.} \label{fig:dataset pipeline}
\end{figure}

Our pipeline for ATE with distant supervision is depicted in Fig.~\ref{fig:dataset pipeline}. We start from raw review corpora (i.e. opinion texts) in order to construct a new dataset for ATE. However, we identify the problem that such corpora contain a lot of noisy sentences, e.g. "I bought this laptop for my kids". Such sentences do not contain opinionated aspect terms and therefore introduce noise to the dataset we wish to build. To overcome this problem, we employ our attention-based method to extract non-noisy sentences from the review corpora and show that sentence selection matters when it comes to constructing a new dataset for ATE.
 
The distant training labels are used as review ratings with an ultimate goal of labelling the tokens of sentences as aspect terms. We exploit the review ratings in a way that assigns a sentiment score on each sentence of the review. To achieve that, we leverage the attention mechanism of a model trained to predict the review ratings (e.g. 1-5 stars) of a review text. Then, we select sentences with high sentiment score since we consider that they are likely to contain domain-related opinionated aspect terms. Finally, we automatically label the tokens of the selected sentences by specifying which tokens are aspect terms.

We use the automatically labelled dataset, that we construct, in order to train a classifier for ATE. We highlight that the resulting token-labelled dataset contains information from the distant training labels (e.g. review labels) and therefore can be used for ATE with distant supervision. 

We perform experiments to show that 
\begin{inlinelist}
    \item ATE with distant supervision can outperform unsupervised systems and rule-based ATE methods and
    \item sentence selection matters while building new ATE datasets.
\end{inlinelist}
To this end, we evaluate our ATE classifier on the human annotated datasets of the SemEval-2014 ABSA contest. Our results show that distantly-supervised ATE achieves a performance higher than unsupervised systems and rule-based baselines~\cite{semeval2014}. Last but not least, our experiments reveal that training a classifier on a set of selected sentences leads to higher ATE performance compared to using the whole sentence set, i.e. sentence selection matters.

The rest of this paper is organized as follows. Section~\ref{sec:related work} presents the related work for ATE and models with attention mechanism. Our sentence selection method is described in Sections~\ref{sec:sentence selection}. Section~\ref{sec:data labelling} analyzes our automated data labelling process. We conduct experiments and present results for ATE with distant supervision in Section~\ref{sec:experiments}. Finally, our work is concluded in Section~\ref{sec:conclusion}.

\section{Related Work} \label{sec:related work}
Research in the area of both supervised and unsupervised ATE has thrived since the first SemEval ABSA task in 2014. Participants who work on supervised ATE~\cite{winners2014, nlang2015, ihs} use the provided human annotated datasets in order to extract features. These features are very similar to those used in traditional Name Entity Recognition (NER) systems~\cite{nerFeatures}. Moreover, participants exploit external sources, such as the WordNet files~\cite{wordnet} and word clusters (e.g. Brown clusters~\cite{brown}). Finally, they usually exploit gazetteers~\cite{gazetteers} and word embeddings~\cite{we}. The extracted features are used to train a classifier such as Conditional Random Fields (CRF) or Support Vector Machine (SVM).

Most unsupervised systems for ATE follow rule-based approaches.~\cite{liuRules} uses relational and syntactic rules to automatically detect aspect terms. Authors of~\cite{v3} present a graph-based approach where nouns and adjectives of large corpora are used as nodes in a graph. Then, they create a list with top-ranking nouns and use it to annotate unseen data by performing exact or lemma matching.

Systems similar to~\cite{absaCzech, pathEmbeddings, Poria201642} perform semi-supervised ATE. They start by creating features (e.g. new word embeddings) using large corpora (e.g. available reviews). These features are later used in order to enrich the feature space of human annotated datasets.

Although there are a lot of publications for sentence selection, they mainly focus on summarizing tasks and not on ATE.~\cite{hadano2011aspect} and~\cite{shimada2011multi} investigate sentence clustering in order to acquire new similar sentences to improve their model. The former focus on aspect identification while the latter on multi-aspects review summarization.~\cite{zhu2013graph} explores the use of a community-leader detection problem with sentence selection in order to build better opinion summarization, where communities consist of a cluster of sentences towards the same aspect of an entity. No existing work seems to have investigated the use of attention for sentence selection in combination with ATE.

To the best of our knowledge, we are the first to perform ATE using distant supervision. We start by using a model with attention mechanism which performs sentence selection. Regarding the attention mechanism, our work is similar to~\cite{attention_mechanism}. Nevertheless, we put forward the focus on the sentence level instead of the word level. We think that modeling sentence representations by sentence embeddings would give better results than using a sequence of word hidden states from a bidirectional long short-term memory (B-LSTM) network. Then, we use the selected sentences in order to create an automatically labelled dataset. Finally, we train a model using the automatically labelled dataset and perform ATE using distant supervision.

\section{Dataset Construction via Sentence Selection} \label{sec:sentence selection}

Given a product review, it often happens that only a subset of its sentences are non-noisy and express an opinion about the item. For example, reviews usually start with sentences which do not contain any useful information about the product review, e.g. "I bought this laptop a few weeks ago". Such sentences are unlikely to contain opinionated aspect terms and are therefore not suitable candidates for constructing datasets for ATE.

We hypothesize that sentence selection matters when it comes to constructing datasets for ATE. In other words, we would like to show that training a model on an automatically filtered dataset (i.e. with few noisy sentences) leads to better ATE performance compared to using the whole one.

In order to perform sentence selection, we build a model to predict the rating of a review (e.g. 1-5 stars). During training, the model assigns weights to all sentences in a review. The higher the value of the weight for a particular sentence, the more important this sentence is for the review classification. Based on these weights, we then devise a method to keep the important sentences and filter out the noisy sentences.

\subsection{Rating Prediction for Sentence Selection} \label{ssec:sentence selection model}
\begin{figure*}[]
\begin{center}
\includegraphics[width=0.8\textwidth]{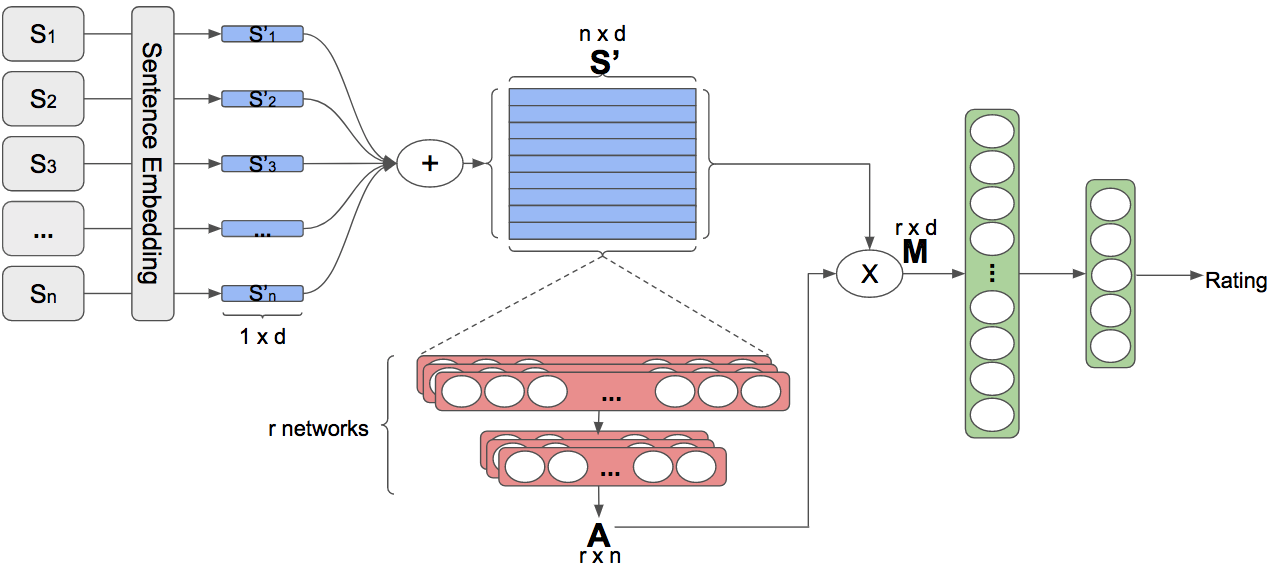}
\end{center}
\caption{Architecture for the review rating predictor. Blue components map to the representation of the review with sentence embeddings. Red components correspond to the $r$~hop attention mechanism. Green components represent the neural network applied on the downstream application, i.e. rating prediction in our case.} \label{fig:attention mechanism}
\end{figure*}

The architecture of the rating prediction model is depicted in Fig.~\ref{fig:attention mechanism} and is inspired by~\cite{attention_mechanism}. However, we differ from~\cite{attention_mechanism} since we force the attention to focus on sentence level rather than on word level. To this end, we remove the B-LSTM layer used in~\cite{attention_mechanism} and feed directly sentence embeddings (instead of word embeddings). The latter are derived using an extension of the Continuous Bag-of-Words Model (CBOW)~\cite{sent2vec}.

We use the publicly available review dataset $D$ from \textit{Amazon}\footnote{\url{http://jmcauley.ucsd.edu/data/amazon/}}. Each review $S~=~(S_1, S_2, ..., S_n)$ consists of $n$~sentences, at most. In order to leverage the attention mechanism, we only keep reviews that have at least $n_{min} = 3$ and at most $n_{max} = 10$ sentences. We pad reviews with $n < n_{max}$ sentences with a special tag so that the length of all reviews equals to $n_{max}$. In that way, we construct a reduced review dataset $D_r$ with a total of $23296$ reviews. We create a train, validation and test set consisting of $20224$, $2048$ and $1024$ reviews respectively. Furthermore, to avoid having most of the attention given to the first sentences, all reviews are shuffled during training. This can be seen as a way of regularization. 

We convert every sentence $S_i$ of each review $S~=~(S_1, S_2, ..., S_{n_{max}}) \in D_r$ to its $d$-dimensional sentence embedding representation $S'_i$. Then, we concatenate all the representations in a $n \times d$ matrix $S'$ (blue components in Fig.~\ref{fig:attention mechanism}). In our case, we use the pre-trained model of~\cite{sent2vec} to compute 600-dimensional sentence embeddings ($d=600$).

For the attention mechanism (red components in Fig.~\ref{fig:attention mechanism}), we adopt the technique and use the same mathematical representation of~\cite{attention_mechanism}. Hence, the $r$-hop attention matrix $A$ of dimensions $r \times n$ is given by
\begin{equation} \label{eq:matrix A}
    A=softmax(W_{a2}~\tanh(W_{a1}{S'}^T))    
\end{equation}

Equation \ref{eq:matrix A} represents a 2-layer MultiLayer Perceptron (MLP) without bias. The hidden layer uses the \textit{tanh} activation function (as recommended by \cite{glorot2010understanding, lecun2012efficient}) and the output layer uses a $softmax$. The weight matrices of the MLP are $W_{a1}$ of dimensions $d_a \times d$ and $W_{a2}$ of dimensions $r \times d_a$, where $d_a$ is a hyperparameter that corresponds to the size of the internal representation.

Similar to~\cite{attention_mechanism}, we compute $r$ weighted sums by multiplying the annotation matrix $A$ and the concatenated sentence embeddings $S'$. The resulting matrix for the sentence embeddings is given by
\begin{equation}
    M=AS'
\end{equation}

Finally, the representation of the review $M$ is fed into another neural network (green components in Fig.~\ref{fig:attention mechanism}). This is necessary because we learn the attention mechanism by minimizing the objective function with respect to a specific task, i.e. review rating prediction in our case. However, we emphasize that the overall model is task-independent.

We innovate by leveraging the attention mechanism of Fig.~\ref{fig:attention mechanism} with the goals of performing sentence selection and constructing a new dataset for ATE. To this end, we modify the architecture of~\cite{attention_mechanism} in order to force the attention mechanism to focus on sentence rather than token level. We highlight that our goal is neither to perform nor to improve review classification and therefore our work is not comparable to~\cite{attention_mechanism}.

\subsection{Attention-based Sentence Filtering} \label{ssec:sentence selection method}
We intend to use the learned weights of the attention mechanism ($W_{a1}$ and $W_{a2}$) in order to perform sentence selection, i.e. filter out noisy sentences. Once the model is trained, we feed once again all reviews without any shuffling. The $A$ matrix contains an attention of $r$~hops for each one of the $n$ sentences in a review. Our goal is to end up with a scalar weight per sentence which indicates the importance of it during the review classification process.

We sum up over all $r$ annotation vectors $a_i$ of matrix $A$ and come up with a vector $\mathbf{\bar{a}}$ of size $n$. Then, we normalize $\mathbf{\bar{a}}$ so that it has a minimum value of $0$ and a maximum value of $1$. Each element $x_i$ of the normalized $\mathbf{\bar{a}}$ maps to a weight for the $i$-th sentence in the review. In turn, the value of $x_i$ reflects the importance of each sentence for the review rating prediction task. The less important the sentence, the lower the value of $x_i$. Hence, the normalized $\mathbf{\bar{a}}$ vector gives a general view of the importance level of all sentences.

\begin{figure}[]
\begin{center}
\includegraphics[width=.75\linewidth]{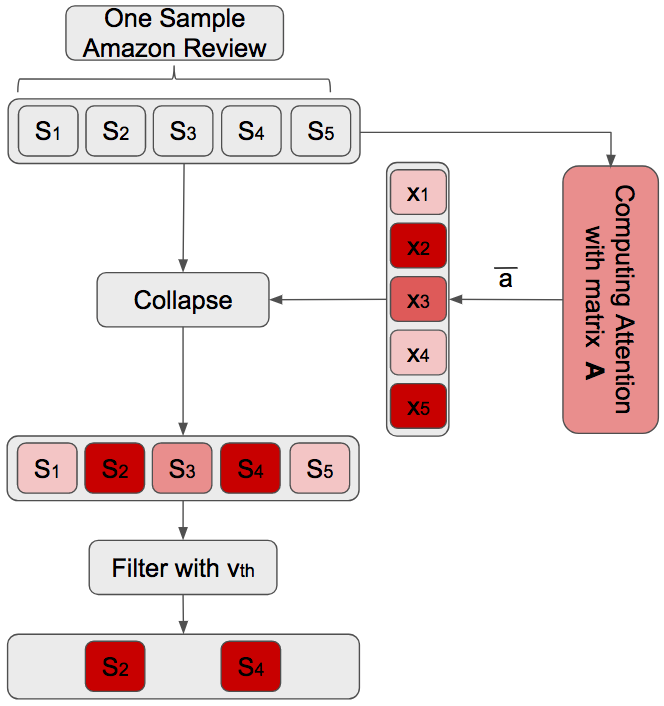}
\end{center}
\caption{Pipeline for the sentence selection, using the attention matrix $A$ learned via review rating prediction as shown in Fig.~\ref{fig:attention mechanism}. The darker the color, the larger the attention weight of the sentence in the review.} \label{fig:pipeline}
\end{figure}

We perform sentence selection by exploiting the elements $x_i$ of the normalized $\mathbf{\bar{a}}$. We consider that sentences with low attention scores do not carry important information and are therefore unlikely to contain opinionated aspect terms. Therefore, we remove sentences with attention score lower than a threshold $v_{th}$. The sentence selection method can be visualized in Fig.~\ref{fig:pipeline} and is described in Algorithm~\ref{alg:sentence selection}. 

\begin{algorithm}
\caption{Sentence Selection} \label{alg:sentence selection}
\begin{algorithmic}[1]
\Require $v_{th} \in [0,1]$, $S'$, $W_{a1}$, $W_{a2}$, review\_sentence\_set
\State $A=softmax(W_{a2}~\tanh(W_{a1}{S'}^T)) $
\State $\mathbf{\bar{a}}$ = convert\_matrix($A$)
\For{weight \textbf{in} $\mathbf{\bar{a}}$}
    \If {weight $<$ $v_{th}$}
        \State p = get\_position(weight, $\mathbf{\bar{a}}$)
        \State remove\_sentence(p, review\_sentence\_set)
    \EndIf    
\EndFor
\State \Return review\_sentence\_set
\end{algorithmic}
\end{algorithm}

\section{Automated Data Labelling} \label{sec:data labelling}
Now that we have a cleaner dataset, we need to label it, in order to use it to train a classifier for ATE. First we explain or automated labelling method. Then we give a visual example of the sentence selection and the labeling for better illustration.

\subsection{Automated Labelling Method} \label{ssec:labelling method}

We use the important review sentences obtained in Section~\ref{ssec:sentence selection method} in order to construct an automatically labelled dataset for ATE. To this end, we label each token of the unlabelled opinion texts in an automated way using the IOB format (short for Inside, Out and Begin) \cite{Poria201642}. Tokens that are aspect terms are labelled with B. In case an aspect term consists of multiple tokens, the first token receives the B label and the rest receive the I label. Tokens that are not aspect terms are labelled with O. The automated data labelling is done using the method described in~\cite{ourpaper}.

We use the following assumptions and tools in order to construct the automatically labelled dataset for ATE.
\begin{itemize}
    \item We only consider nouns and noun phrases~\cite{liuRules} as candidate aspect terms. However, we use nouns and noun phrases that appear less than 30 times (i.e. minimum support 30) in the clean dataset $D_r$. This reduces the noise in the aspect term labelling introduced by infrequent nouns and noun phrases.
    \item Aspect terms are often objects of verbs ("I like this laptop") or are accompanied by modifiers (e.g. "The screen is perfect") that express a sentiment. Hence, we use the Senticnet sentiment lexicon~\cite{senticnet_lexicon} in order to check if words that describe candidate aspect terms express a sentiment or not.
    \item We exploit a set of 12 syntactic rules that are able to capture aspect terms. These rules check if there are syntactic dependencies between opinionated adjectives or verbs and nouns or noun phrases. A subset of these rules is tabulated in Table~\ref{tab:syntactic rules}. For the syntactic rules, we adopt a notation similar to~\cite{ourpaper}.    
\end{itemize}

\begin{table*}[]
\centering
\caption{Subset of syntactic rules for aspect term extraction.}
\label{tab:syntactic rules}
\resizebox{\textwidth}{!}{
\begin{tabular}{l|c|c}
\multicolumn{1}{c|}{\textbf{Rules}} & \textbf{Example} & \multicolumn{1}{c}{\textbf{Extracted Targets}} \\ \hline \hline
\begin{tabular}[c]{@{}l@{}} $depends(dobj, t_i, t_j)$ \textbf{and} $opinion\_word(t_j)$ \\ \textbf{then} $mark\_target(t_i)$ \end{tabular} & I love this laptop & \multicolumn{1}{c}{laptop} \\ \hline
\begin{tabular}[c]{@{}l@{}} $depends(nsubj, t_i, t_j)$ \textbf{and} $depends(acomp, t_k, t_j)$ \\  \textbf{and} $opinion\_word(t_k)$ \textbf{then} $mark\_target(t_i)$ \end{tabular} & The GPU is perfect & GPU \\ \hline
\begin{tabular}[c]{@{}l@{}} $depends(cc~\mathbf{or}~conj, t_i, t_j)$ \textbf{and} $is\_aspect(t_j)$ \\ \textbf{then} $mark\_target(t_i)$ \end{tabular} & Keyboard and sound are awful & \begin{tabular}[c]{@{}c@{}}keyboard \\ sound\end{tabular} \\ \hline
\begin{tabular}[c]{@{}l@{}} $depends(compound, t_i, t_j)$ \textbf{and} $is\_aspect(t_j)$ \\ \textbf{then} $mark\_target(t_i)$ \end{tabular} & The retina display is superb & retina display 
\end{tabular}
}
\end{table*}
The functions of Table~\ref{tab:syntactic rules} can be interpreted as follows:
\begin{itemize}[topsep=1pt, itemsep=-1ex, partopsep=1ex, parsep=1ex]
    \item $depends(d,t_i,t_j)$ is true if the syntactic dependency between the tokens $t_i$ and $t_j$ is $d$.
    \item $opinion\_word(t_i)$ is true if the token $t_i$ is in the sentiment lexicon.
    \item $mark\_target(t_i)$ means that we mark the token $t_i$ as aspect term.
    \item $is\_aspect(t_i)$ is true if the token $t_i$ is already marked as aspect term.
\end{itemize}

Algorithm~\ref{alg:labelling} describes the automated method we use in order to annotate the tokens of the non-noisy sentences with the IOB format. This algorithm is similar to~\cite{ourpaper}. However,~\cite{ourpaper} focus on achieving high precision values for ATE. Since we are interested in the \textit{F-score}, we apply some modifications on~\cite{ourpaper}. To this end, we use a bigger set of syntactic rules, a bigger sentiment lexicon and remove the list of quality phrases.

\begin{algorithm}
\caption{Automated Data Labelling} \label{alg:labelling}
\begin{algorithmic}[1]
\Require corpus of filtered sentences, set of syntactic rules, sentiment lexicon
\For{sentence \textbf{in} corpus}
    \State labels = []
        \For{token \textbf{in} sentence}
            \If {token \textbf{is} NOUN}
                \State \emph{l = get\_label(token, rules, lexicon)}
                \State \emph{labels.append(l)}
            \EndIf
        \EndFor
    \State \emph{assign\_iob\_tags(sentence, labels)}
\EndFor
\end{algorithmic}
\end{algorithm}


It is obvious that the resulting dataset carries some information from training signals (i.e. review ratings) not directly related to the token-based labelling. 
Therefore, we use the automatically labelled dataset to perform ATE with distant supervision.

\subsection{Data Labelling Visualization} \label{ssec:visualization}
\begin{figure}
    \centering
    \includegraphics[width=1\linewidth]{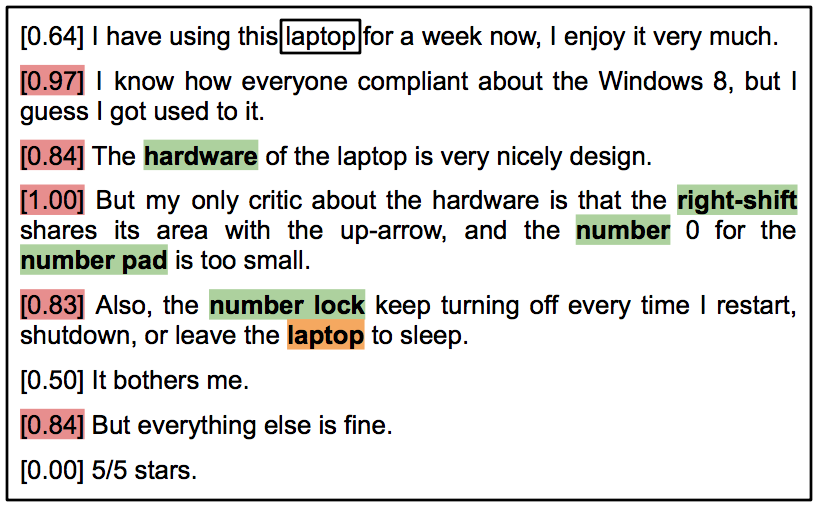}
    \caption{Annotated sample review with 5 stars from the Amazon dataset. Each sentence of the review is shown with its attention score. Extracted aspect terms are highlighted in green and orange, where the first color represents true aspects and the second noisy ones.}
    \label{fig:annotated_samples}
\end{figure}

Figure~\ref{fig:annotated_samples} depicts the results of our automated data labelling process applied on a 5-star review. As we can see, sentences with strong opinions are more highlighted by the attention mechanism and those which do not carry important information are less pointed out. However, these do not have necessarily no attention because they are still relevant for the task of the review rating prediction. Moreover, we observe that our simple regularization method to avoid having all the attentions focused on the first sentences works, i.e. attention can be given to any sentences in the review.

For the example of Fig.~\ref{fig:annotated_samples}, we filter out sentences with an attention score $v_{th} < 0.7$ as described in Section~\ref{sec:sentence selection}. For the remaining sentences (highlighted in red in Fig.~\ref{fig:annotated_samples}), we apply the automated data labelling process. Tokens that are nouns and obey at least one of the syntactic rules are marked as aspect terms (depicted in green in Fig.~\ref{fig:annotated_samples}). 

With our automated dataset construction we try to optimize the quality of the extracted aspect terms. We achieve that by annotating selected sentences using a set of syntactic rules and a sentiment lexicon. The sentence selection method removes irrelevant or noisy aspect terms (e.g. "laptop" in the first sentence of Fig.~\ref{fig:annotated_samples}). However, some remaining sentences might still contain noisy aspects as depicted in orange in Fig.~\ref{fig:annotated_samples}.

The data labelling visualization verifies that our assumptions (Section~\ref{ssec:labelling method}) are correct. More concretely, we see that our automated data labelling method manages to detect and label successfully aspect terms in the sentences by exploiting nouns and noun phrases, syntactic rules and a sentiment lexicon.

\section{Experiments} \label{sec:experiments}
We perform ATE with distant supervision by training a model using the automatically labelled dataset (hereafter denoted as ALD) as training data. We aim at measuring the effectiveness of our proposed sentence selection mechanism (Section~\ref{sec:sentence selection}) and prove our hypothesis that sentence selection matters when it comes to constructing a new dataset for ATE. We evaluate our classifier using the human labelled test dataset (hereafter denoted as HLD) of the SemEval-2014 ABSA contest~\cite{semeval2014} for the laptop domain. As evaluation metric, we use the CoNLL\footnote{\url{http://www.cnts.ua.ac.be/conll2003/}} \textit{F-score} which is given by Eq.~\ref{eq:F-score}.

\begin{equation} \label{eq:F-score}
    F_1 = \frac{2 \cdot P \cdot R}{P+R}
\end{equation}
$P$ and $R$ stand for precision and recall and are given by Eq.~\ref{eq:precision} and~\ref{eq:recall} respectively. $S$ and $G$ are the sets of retrieved and correct aspect terms of our system respectively.
\begin{multicols}{2}
    \noindent
    \begin{equation} \label{eq:precision}
        P = \frac{|S \cap G|}{|S|}
    \end{equation}
    \begin{equation} \label{eq:recall}
        R = \frac{|S \cap G|}{|G|}
    \end{equation} 
\end{multicols}
\noindent

We start by building a series of baseline models in order to prove that the syntactic rules we use are capable of extracting aspect terms. Then, we use the ALD to train a model and evaluate it on the HLD. The results show that the trained classifier outperforms the baselines models and validates our hypothesis, i.e. sentence selection matters.

\subsection{Rule-based Baselines}

Unsupervised ATE systems like \cite{v3, spanishPaper} use syntactic rules in order to identify aspect terms. A simple rule-based model may identify as aspects nouns with any syntactic dependency to any word with positive or negative polarity (e.g. "good", "bad"). More sophisticated rule-based systems \cite{liuRules} capture aspect terms by linking nouns to modifiers and adjectival complements. We wish to prove that the syntactic rules we use, combined with the Senticnet lexicon, are capable of extracting aspect terms. 

We create 4 different rule-based baseline models that do no use any machine learning algorithm. Each baseline is more advanced than the previous one, since it exploits a bigger set of syntactic rules. During the prediction process, a token of the HLD is labelled as a target if 
\begin{inlinelist}
    \item it is a noun and
    \item satisfies at least one of the syntactic rules of each baseline.
\end{inlinelist}

\begin{itemize}
    \item \textbf{Baseline B-1:} It considers as aspect terms all nouns of the test set with any syntactic relation to any word from the Senticnet lexicon.
    \item \textbf{Baseline B-2:} It is similar to B-1, however nouns are considered as aspect terms only if they have any syntactic relation to adjectives from the Senticnet lexicon.
    \item \textbf{Baseline B-3:} It extends B-2 by labelling nouns as aspect terms only if the syntactic relation to any word from the Senticnet lexicon is of type \textit{amod} or \textit{advmod} or \textit{acomp}. Moreover, B-3 labels as aspect terms nouns that are in conjunction with other aspect terms.
    \item \textbf{Baseline B-4:} It includes the full set of the 12 syntactic rules we introduce. Once again, only nouns related to words from the Senticnet lexicon are considered as aspect terms.
\end{itemize}
Results for the baseline models are tabulated in Table~\ref{tab:experimental results}. We see that B-4 performs the best among all baselines. This fact proves that our set of syntactic rules, combined with the Senticnet lexicon, are capable of identifying aspect terms. We also highlight that these baselines are completely unsupervised and domain-independent since they use only syntactic rules combined with a sentiment lexicon in order to identify aspect terms, i.e there is no use of training data or attention.

\begin{table*}[]
\centering
\caption{Experimental results (precision, recall and \textit{F-score}) for ATE using distant supervision. The labels of the columns indicate the model used for ATE. In case of the SVM classifier, the subscript indicates the value used for sentence selection.} \label{tab:experimental results}
\resizebox{\textwidth}{!}{%
\begin{tabular}{c||c|c|c|c||c|c|c|c|c|c||c||c}
 & {\rotatebox[origin=c]{90}{B-1}} & {\rotatebox[origin=c]{90}{B-2}} & {\rotatebox[origin=c]{90}{B-3}} & {\rotatebox[origin=c]{90}{B-4}} & {\rotatebox[origin=c]{90}{SVM$_{0}$}} & {\rotatebox[origin=c]{90}{SVM$_{0.5}$}} & {\rotatebox[origin=c]{90}{SVM$_{0.6}$}} & {\rotatebox[origin=c]{90}{SVM$_{0.7}$}} & {\rotatebox[origin=c]{90}{SVM$_{0.8}$}} & {\rotatebox[origin=c]{90}{SVM$_{0.99}$}} & {\rotatebox[origin=c]{90}{\begin{tabular}[c]{@{}c@{}} SVM$_{0.7}$ \\ with np\end{tabular}}} & {\rotatebox[origin=c]{90}{\begin{tabular}[c]{@{}c@{}}~B-LSTM \\ \& CRF\end{tabular}}} \\ \hline \hline
$\mathbf{P}$ & 21.01 & 30.45 & 32.44 & 40.22 & 47.97 & 48.43 & 47.84 & 48.81 & 47.82 & 46.09 & 47.95 & \textbf{50.33}  \\ \hline
$\mathbf{R}$ & 16.34 & 11.20 & 16.49 & 33.28 & 36.35 & 37.47 & 37.19 & 38.12 & 36.81 & 35.14 & 40.26 & \textbf{40.49} \\ \hline
$\mathbf{F_1}$ & 18.38 & 16.37 & 21.87 & 36.42 & 41.36 & 42.25 & 41.85 & 42.80 & 41.60 & 39.87 & 43.77 & \textbf{44.87}
\end{tabular}
}
\end{table*}

\subsection{SVM for Aspect Term Extraction}
We wish to beat the baseline models by using machine learning. We start by defining 6 different thresholds $v_{th} \in [0, 0.5, 0.6, 0.7, 0.8, 0.99]$ and perform sentence selection using the method of Section~\ref{sec:sentence selection} with the hyperparameters $r = 30$ and $d_a = 350$. This results in 6 different ALDs. Then, we use these datasets |~one at a time~| in order to train an SVM classifier. The classifier is evaluated on the HLD using the CoNLL \textit{F-score}.

We construct baseline features~\cite{stratos} in order to train the SVM\footnote{We use the implementation of \texttt{LIBLINEAR}~\cite{liblinear}.} classifier. More concretely, we build one-hot vectors using the sentence structure. For each token $x_i$ in a sentence, features are created using the identities (string representation) of $x_{i-2}$, $x_{i-1}$, $x_{i}$, $x_{i+1}$ and $x_{i+2}$. In case $x_i$ is at the beginning or at the end of a sentence, special characters (e.g. $\_$s$\_$ and $\_$e$\_$) are used to indicate the start and the end of the sentence respectively. In addition, we build features using the word morphology. For each token $x_i$ in a sentence, we create extra features by taking the prefix and the suffix (up to a length of 4) of $x_i$. Moreover, morphological features are enriched by investigating if $x_i$ is
\begin{inlinelist}
    \item capitalized,
    \item non-alphanumeric or
    \item numeric.
\end{inlinelist}

The performance of the SVM classifier is tabulated in Table~\ref{tab:experimental results}. Columns SVM$_{0}$ through SVM$_{0.99}$ prove that the SVM classifier beats the baseline model. The subscript in the column name indicates the value of $v_{th}$. These columns also validate our hypothesis, that sentence selection matters when it comes to constructing a new dataset for ATE. The classifier has a performance of $F_1=41.36$ when we use all sentences ($v_{th} = 0$) for the ALD construction. All evaluation metrics increase as the sentence selection threshold increases from $0$ to $0.7$, apart from a small fluctuation (0.4\%) when $v_{th}=0.6$. We believe that this increase is due to the fact that the sentence selection removes noise from the ALD which leads to improved classifier performance. We also notice a decrease in the performance for sentence selection thresholds greater than $0.7$. In these cases, we believe that the sentence selection is harsh and removes useful information from the ALD which results in performance deterioration.

We wish to further boost the \textit{F-score} of the SVM classifier. To this end, we use $v_{th} = 0.7$ |~since this threshold gives the best performance so far~| and build a new ALD by considering nouns and noun phrases (np) as candidate aspect terms. In that way, we improve the performance of the classifier from  $F_1=42.80$ to $F_1=43.77$.

\subsection{B-LSTM \& CRF for Aspect Term Extraction}
\begin{figure}[]
    \begin{center}
    \includegraphics[width=0.47\textwidth]{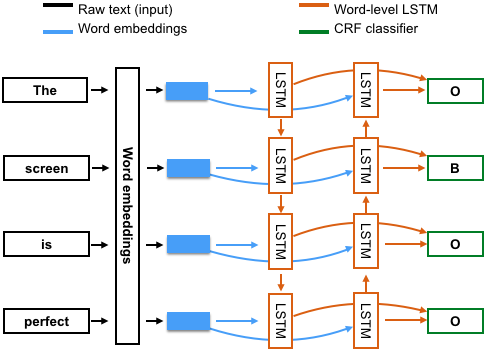}
    \end{center}
    \caption{Architecture of the B-LSTM \& CRF model. The features extracted from the B-LSTM layer are used by the CRF for sequential labelling.} \label{fig:neuroate}
\end{figure}

We exploit an architecture that employs a B-LSTM followed by a CRF classifier in order to further boost the performance for ATE using distant supervision. To this end, we choose the ALD constructed with $v_{th}=0.7$ and noun phrases, that gives the best performance, and train a B-LSTM \& CRF classifier (Fig.~\ref{fig:neuroate}). Then, we evaluate our model on the HLD, i.e. the human annotated test set of the SemEval-2014 ABSA task. We also use the training set of the SemEval-2014 ABSA task as a validation set.

The B-LSTM layer of the B-LSTM \& CRF classifier exploits the structure of the sentence (i.e. previous and next words of each token) in order to extract new features (depicted in orange in Fig.~\ref{fig:neuroate}). These features are given as input to the CRF classifier, which performs sequential labelling.

In order to train the model, we use the 300-dimensional pre-trained word embeddings of fastText\footnote{https://github.com/facebookresearch/fastText}. Moreover, we use 100 hidden states for each LSTM cell. The classifier is trained for maximum 20 epochs and uses a patience value of 5, i.e. the training terminates if there is no improvement on the validation set for more than 5 consecutive epochs. Finally, we use the Adam optimizer~\cite{DBLP:journals/corr/KingmaB14} with learning rate $\alpha=0.01$ and a batch size of 32.

We perform 25 experiments in order to report mean values for the precision, recall and $F_1$. We also construct 95\% confidence intervals for the aforementioned metrics. The obtained results for precision, recall and \textit{F-score} are:
\begin{inlinelist}
    \item $P = 50.33 \pm 0.38$,
    \item $R = 40.49 \pm 0.55$ and
    \item $F_1 = 44.87 \pm 0.42$.
\end{inlinelist}
The experimental results validate that the B-LSTM \& CRF classifier outperforms all aforementioned models for all 3 evaluation metrics. The mean values are tabulated in Table~\ref{tab:experimental results}.

Last but not least, our experimental results for ATE with distant supervision reveal that our method outperforms the supervised baseline method of the SemEval-2014 ABSA contest. More concretely, our method achieves an \textit{F-score} of $44.87$ compared to the supervised baseline \textit{F-score} of $35.64$, i.e. a relative improvement of $25.9\%$.

In this work we mainly focus on proving that sentence selection matters when it comes to constructing a new dataset for ATE. We leave the experimentation with various deep learning architectures \cite{recent_trends} and comparison against state-of-the-art models for future work.

\section{Conclusion} \label{sec:conclusion}
In this paper, we first show that sentence selection matters when it comes to building a corpus for ATE. We start from publicly available review corpora and exploit a multi-hop and task-independent attention mechanism. We force this mechanism to focus on sentence level, i.e. to give an attention score to each sentence of a review. We then perform sentence selection by varying the attention threshold from 0 to 1. 

Secondly, we annotate the tokens of the selected sentences and construct new datasets for ATE~|~one for each attention threshold. To this end, we employ our automated data labelling method. We train multiple classifiers using the automatically labelled datasets and evaluate them on the human labelled dataset of the SemEval-2014 ABSA contest. We observe that all evaluation metrics behave similarly to an inverted U-shaped curve as the sentence selection threshold increases. 

Our experiments validate our hypothesis that sentence selection matters when it comes to constructing a new dataset for ATE. Moreover, we show that ATE with distant supervision outperforms all our unsupervised rule-based models, as well as the supervised baseline of SemEval-2014 ABSA task.


\bibliography{bibliography}
\bibliographystyle{IEEEtran}

\end{document}